\documentclass[11pt, a4paper, logo, copyright, nonumbering]{deepseek}
\usepackage[authoryear, sort&compress, round]{natbib}
\usepackage{dblfloatfix}
\usepackage{ulem}
\usepackage{caption}
\usepackage{dramatist}
\usepackage{xspace}
\usepackage{pifont}
\usepackage{multirow}
\usepackage{tcolorbox}
\usepackage{xltabular}
\usepackage{longtable}
\usepackage{hyperref}
\usepackage[capitalize,noabbrev]{cleveref} 
\usepackage{amsfonts}
\usepackage{amsmath}
\usepackage{amssymb}
\usepackage{lineno}

\usepackage[bottom]{footmisc}
\usepackage{CJKutf8}
\usepackage{subfigure}
\usepackage{setspace}

\makeatletter
\def\@BTrule[#1]{%
  \ifx\longtable\undefined
    \let\@BTswitch\@BTnormal
  \else\ifx\hline\LT@hline
    \nobreak
    \let\@BTswitch\@BLTrule
  \else
     \let\@BTswitch\@BTnormal
  \fi\fi
  \global\@thisrulewidth=#1\relax
  \ifnum\@thisruleclass=\tw@\vskip\@aboverulesep\else
  \ifnum\@lastruleclass=\z@\vskip\@aboverulesep\else
  \ifnum\@lastruleclass=\@ne\vskip\doublerulesep\fi\fi\fi
  \@BTswitch}
\makeatother

\addto\extrasenglish{
}

 {\begin{list}{}%
         {\setlength{\leftmargin}{#1}}%
         \item[]%
 }
 {\end{list}}
 
\reportnumber{001} % Leave blank if n/a

\renewcommand{\today}{}

\title{\centering Native Sparse Attention: Hardware-Aligned and Natively Trainable Sparse Attention}

\author[*]{
\small
Jingyang Yuan$^{*1,2}$,
Huazuo Gao$^{1}$,
Damai Dai$^{1}$,
Junyu Luo$^{2}$,
Liang Zhao$^{1}$,
Zhengyan Zhang$^{1}$,
Zhenda Xie$^{1}$,
Y. X. Wei$^{1}$,
Lean Wang$^{1}$,
Zhiping Xiao$^{3}$,
Yuqing Wang$^{1}$,
Chong Ruan$^{1}$,
Ming Zhang$^{2}$,
Wenfeng Liang$^{1}$,
Wangding Zeng$^{1}$
\\
\small
$^1$DeepSeek-AI \\
\small
$^2$Key Laboratory for Multimedia Information Processing, School of Computer Science, Peking University, PKU-Anker LLM Lab \\
\small
$^3$University of Washington \\
\small
\texttt{\{yuanjy, mzhang\_cs\}@pku.edu.cn},
\small
\texttt{\{zengwangding, wenfeng.liang\}@deepseek.com}
\\
\small
}
\correspondingauthor{Contribution during internship at DeepSeek-AI.}

\renewcommand{\phi}{\varphi}

\renewcommand{\leq}{\leqslant}

\renewcommand{\epsilon}{\varepsilon}
\renewcommand{\imath}{\mathrm{i}}

\newlength{\restsubwidth}
\newlength{\restsubheight}
\newlength{\restsubmoreheight}
\newcommand{\rest}[2]{%
        \settowidth{\restsubwidth}{\ensuremath{#2}}
        \settoheight{\restsubheight}{\ensuremath{{}_{#2}}}
        \ensuremath{{#1\hskip 0.5pt}_{\vrule\kern2pt\parbox[b][%
        4pt][b]{\the\restsubwidth}{%
                        \ensuremath{{}_{#2}}}}}
        }

\def\method{NSA}

\begin{abstract}

Long-context modeling is crucial for next-generation language models, yet the high computational cost of standard attention mechanisms poses significant computational challenges. 
Sparse attention offers a promising direction for improving efficiency while maintaining model capabilities.
We present \method{}, a \underline{N}atively trainable \underline{S}parse \underline{A}ttention mechanism that integrates algorithmic innovations with hardware-aligned optimizations to achieve efficient long-context modeling.
\method{} employs a dynamic hierarchical sparse strategy, combining coarse-grained token compression with fine-grained token selection to preserve both global context awareness and local precision. Our approach advances sparse attention design with two key innovations: (1) We achieve substantial speedups through arithmetic intensity-balanced algorithm design, with implementation optimizations for modern hardware. (2) We enable end-to-end training, reducing pretraining computation without sacrificing model performance.
As shown in \cref{fig:fig1}, experiments
show the model pretrained with \method{} maintains or exceeds Full Attention models
across general benchmarks, long-context tasks, and instruction-based reasoning.
Meanwhile, \method{} achieves substantial speedups over Full Attention on 64k-length sequences across decoding, forward propagation, and backward propagation, validating its efficiency throughout the model lifecycle.

\end{abstract}

\begin{document}
\begin{CJK*}{UTF8}{gbsn}

\maketitle

\section{Introduction}

\begin{figure}[t] 
\centering 
\includegraphics[width=0.7\textwidth]{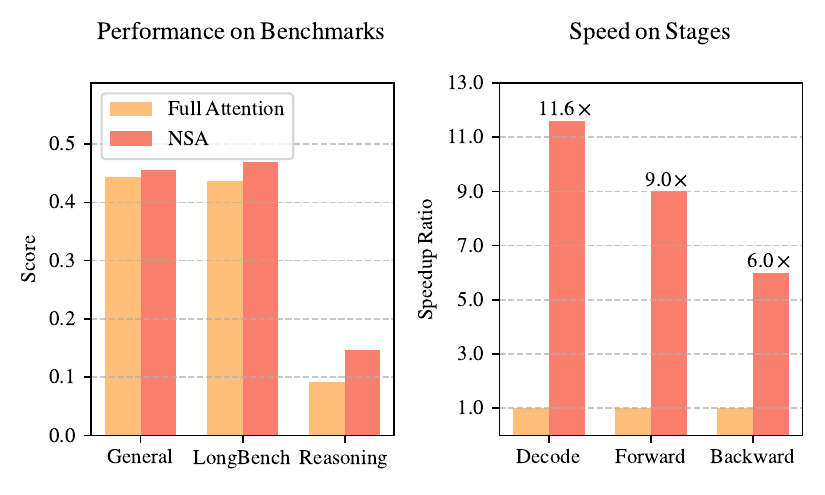} 
\caption{Comparison of performance and efficiency between Full Attention model and our \method{}. Left: Despite being sparse, \method{} surpasses Full Attention baseline on average across general benchmarks, long-context tasks, and reasoning evaluation. Right: For 64k-length sequence processing, \method{} achieves substantial computational speedup compared to Full Attention in all stages: decoding, forward propagation, and backward propagation.
}
\label{fig:fig1}
\end{figure}

The research community increasingly recognizes long-context modeling as a crucial capability for next-generation large language models, driven by diverse real-world applications ranging from in-depth reasoning~\citep{star,deepseekr1}, repository-level code generation~\citep{RepoCoder,CodeAgent} and multi-turn autonomous agent systems~\citep{AutoAgent}.
Recent breakthroughs, including OpenAI's o-series models, DeepSeek-R1~\citep{deepseekr1}, and Gemini 1.5 Pro~\citep{team2024gemini}, 
enabling models to process entire codebases, lengthy documents, maintain coherent multi-turn conversations over thousands of tokens, and perform complex reasoning across long-range dependencies.
However, the high complexity~\citep{zaheer2020big} of vanilla Attention~\citep{vaswani2017attention} mechanisms emerges as a critical latency bottleneck as sequence length increases. Theoretical estimates indicate that attention computation with softmax architectures accounts for 70--80\% of total latency when decoding 64k-length contexts, underscoring the urgent need for more efficient attention mechanisms.

A natural approach to efficient long-context modeling is to take advantage of the inherent sparsity of softmax attention \citep{ge2023model, jiang2023llmlingua}, where selectively computing critical query-key pairs can significantly reduce computational overhead while preserving performance. Recent advances demonstrate this potential through diverse strategies: 
KV-cache eviction methods~\citep{h2o,snapkv,zhou2024llm}, blockwise KV-cache selection methods~\citep{quest,infllm,gao2024seerattention}, and sampling, clustering or hashing-based selection methods~\citep{magicpig,clusterkv,desai2024hashattention}.
Despite these promising strategies, existing sparse attention methods often fall short in practical deployments. Many approaches fail to achieve speedups comparable to their theoretical gains; moreover, most methods lack effective training-time support to fully exploit the sparsity patterns of attention.

To address these limitations, the deployment of effective sparse attention must tackle two key challenges: (1) \textbf{\textit{Hardware-aligned inference speedup}}: Converting theoretical computation reductions into actual speed improvements
requires hardware-friendly algorithm design during both prefilling and decoding stages to mitigate memory access and hardware scheduling bottlenecks;
(2) \textbf{\textit{Training-aware algorithm design}}: Enabling end-to-end computation with trainable operators to 
reduce training costs while maintaining model performance.
These requirements are crucial for real-world applications to achieve fast long-context inference or training.
When considering both aspects, existing methods still exhibit a noticeable gap.

\begin{figure*}[t] 
\centering 
\includegraphics[width=1\textwidth]{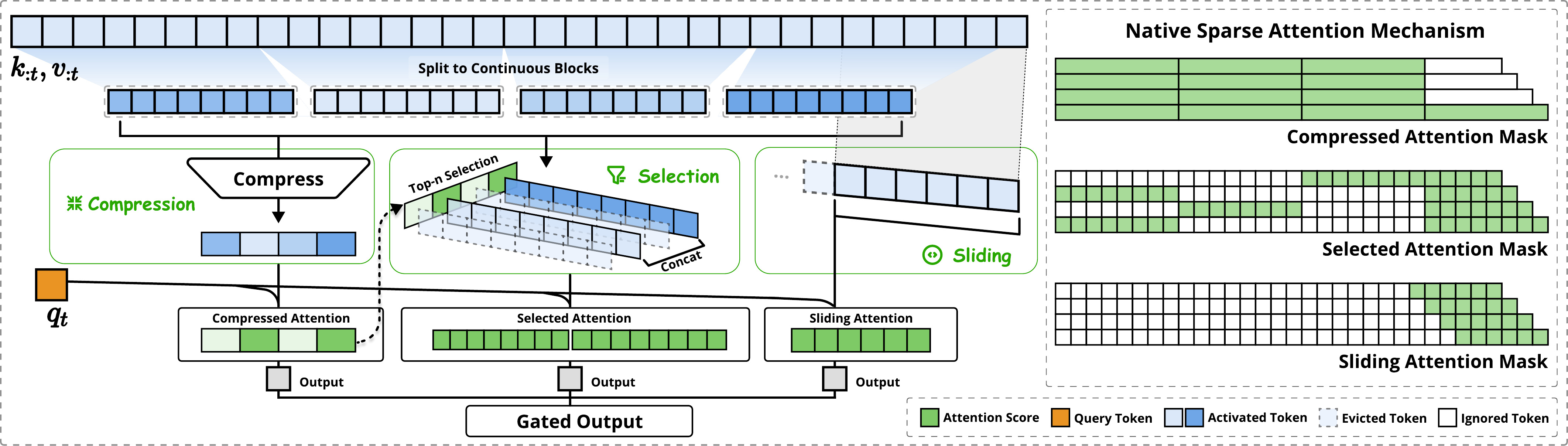} 
\caption{Overview of \method{}'s architecture. Left: The framework processes input sequences through three parallel attention branches: For a given query, preceding keys and values are processed into compressed attention for coarse-grained patterns, selected attention for important token blocks, and sliding attention for local context. Right: Visualization of different attention patterns produced by each branch. Green areas indicate regions where attention scores need to be computed, while white areas represent regions that can be skipped.}
\label{fig:framework}
\end{figure*}

To achieve more effective and efficient sparse attention, we present \method{}, a Natively trainable Sparse Attention architecture that integrates hierarchical token modeling. As shown in \cref{fig:framework}, \method{} reduces per-query computation by organizing keys and values into temporal blocks and processing them through three attention paths: compressed coarse-grained tokens, selectively retained fine-grained tokens, and sliding windows for local contextual information. Then we implement specialized kernels to maximize its practical efficiency.
\method{} introduces two core innovations corresponding to the key requirements above:
(1) Hardware-aligned system: Optimize blockwise sparse attention for Tensor Core utilization and memory access, ensuring balanced arithmetic intensity.
(2) Training-aware design: Enable stable end-to-end training through efficient algorithms and backward operators.
This optimization enables \method{} to support both efficient deployment and end-to-end training.

We evaluate \method{} through comprehensive experiments on real-world language corpora. Pretraining on a 27B-parameter transformer backbone with 260B tokens, we assess \method{}'s performance across general language evaluations, long-context evaluations, and chain-of-thought reasoning evaluation. We further compare the kernel speed on A100 GPUs with optimized Triton~\citep{tillet2019triton} implementations. Experimental results demonstrate that \method{} achieves comparable or superior performance to full attention baseline, while outperforming existing sparse attention approaches. Additionally, \method{} delivers substantial speedups across decoding, forward, and backward stages compared to Full Attention, with the speedup ratio increasing for longer sequences. These results validate that our hierarchical sparse attention design effectively balances model capability and computational efficiency.

\section{Rethinking Sparse Attention Methods}
\label{sec:critique}

Modern sparse attention methods have made significant strides in reducing the theoretical computational complexity of transformer models. However, most approaches predominantly apply sparsity during inference while retaining a pretrained Full Attention backbone, potentially introducing architectural bias that limits their ability to fully exploit sparse attention's advantages. Before introducing our native sparse architecture, we systematically analyze these limitations through two critical lenses.

\subsection{The Illusion of Efficient Inference}

Despite achieving sparsity in attention computation, many methods fail to achieve corresponding reductions in inference latency, primarily due to two challenges:

\textbf{Phase-Restricted Sparsity.}
Methods such as H2O \citep{h2o} apply sparsity during autoregressive decoding while requiring computationally intensive pre-processing (e.g. attention map calculation, index building) during prefilling. In contrast, approaches like MInference \citep{minference} focus solely on prefilling sparsity. 
These methods fail to achieve acceleration across all inference stages, as at least one phase remains computational costs comparable to Full Attention.
The phase specialization reduces the speedup ability of these methods in prefilling-dominated workloads like book summarization and code completion, or decoding-dominated workloads like long chain-of-thought~\citep{cot} reasoning.

\textbf{Incompatibility with Advanced Attention Architecture.}
Some sparse attention methods fail to adapt to modern decoding efficient architectures like Mulitiple-Query Attention~(MQA) \citep{mqa} and Grouped-Query Attention~(GQA) \citep{gqa}, which significantly reduced the memory access bottleneck during decoding by sharing KV across multiple query heads. For instance, in approaches like Quest \citep{quest}, each attention head independently selects its KV-cache subset. Although it demonstrates consistent computation sparsity and memory access sparsity in Multi-Head Attention (MHA) models, it presents a different scenario in models based on architectures like GQA, where the memory access volume of KV-cache corresponds to the union of selections from all query heads within the same GQA group. This architectural characteristic means that while these methods can reduce computation operations, the required KV-cache memory access remains relatively high.
This limitation forces a critical choice: while some sparse attention methods reduce computation, their scattered memory access pattern conflicts with efficient memory access design from advanced architectures.

These limitations arise because many existing sparse attention methods focus on KV-cache reduction or theoretical computation reduction, but struggle to achieve significant latency reduction in advanced frameworks or backends.
This motivates us to develop algorithms that combine both advanced architectural and hardware-efficient implementation to fully leverage sparsity for improving model efficiency.

\subsection{The Myth of Trainable Sparsity}
Our pursuit of native trainable sparse attention is motivated by two key insights from analyzing inference-only approaches:
(1) \textbf{\textit{Performance Degradation}}: Applying sparsity post-hoc forces models to deviate from their pretrained optimization trajectory. As demonstrated by \citet{magicpig}, top 20\% attention can only cover 70\% of the total attention scores, rendering structures like retrieval heads in pretrained models vulnerable to pruning during inference.
(2)~\textbf{\textit{Training Efficiency Demands}}: 
Efficient handling of  long-sequence training is crucial for modern LLM development. This includes both pretraining on longer documents to enhance model capacity, and subsequent adaptation phases such as long-context fine-tuning and reinforcement learning. However, existing sparse attention methods primarily target inference, leaving the computational challenges in training largely unaddressed. This limitation hinders the development of more capable long-context models through efficient training. Additionally, efforts to adapt existing sparse attention for training also expose challenges:

\textbf{Non-Trainable Components.} Discrete operations in methods like ClusterKV~\citep{clusterkv} 
(includes k-means clustering) and MagicPIG~\citep{magicpig} (includes SimHash-based selecting) create discontinuities in the computational graph. These non-trainable components prevent gradient flow through the token selection process, limiting the model's ability to learn optimal sparse patterns. 

\textbf{Inefficient Back-propagation.} Some theoretically trainable sparse attention methods suffer from practical training inefficiencies. Token-granular selection strategy used in approaches like HashAttention~\citep{desai2024hashattention} leads to the need to load a large number of individual tokens from the KV cache during attention computation. 
This non-contiguous memory access prevents efficient adaptation of fast attention techniques like FlashAttention, which rely on contiguous memory access and blockwise computation to achieve high throughput.
As a result, implementations are forced to fall back to low hardware utilization, significantly degrading training efficiency.

\subsection{Native Sparsity as an Imperative}

These limitations in inference efficiency and training viability motivate our fundamental redesign of sparse attention mechanisms.
We propose \method{}, a natively sparse attention framework that addresses both computational efficiency and training requirements.
In the following sections, we detail the algorithmic design and operator implementation of \method{}.

\section{Methodology}

Our technical approach spans algorithm design and kernel optimization.
In the following subsections, we first introduce the background of our methodology. Then we present the overall framework of \method{}, followed by its key algorithmic components. Finally, we detail our hardware-optimized kernel design that maximizes practical efficiency.

\subsection{Background}
\textbf{Attention Mechanism} 
 is widely used in language modeling where each query token \( \mathbf{q}_t \) computes relevance scores against all preceding keys \( \mathbf{k}_{:t} \) to generate a weighted sum of values \( \mathbf{v}_{:t} \). Formally, for an input sequence of length \( t \), the attention operation is defined as:
\begin{equation}
\label{equ:attn}
    \mathbf{o}_t = \operatorname{Attn}\left(\mathbf{q}_t, \mathbf{k}_{:t}, \mathbf{v}_{:t}\right)
\end{equation}
where \( \operatorname{Attn} \) denotes the attention function:
\begin{equation}
    \operatorname{Attn}\left(\mathbf{q}_t, \mathbf{k}_{:t}, \mathbf{v}_{:t}\right) = \sum_{i=1}^t\frac{ \alpha_{t,i} \mathbf{v}_i}{\sum_{j=1}^t \alpha_{t,j}}, \quad \alpha_{t,i} = e^{\frac{\mathbf{q}_t^\top \mathbf{k}_i}{\sqrt{d_k}}}\,.
\end{equation}
Here, \( \alpha_{t,i} \) represents the attention weight between \( \mathbf{q}_t \) and \( \mathbf{k}_i \), and \( d_k \) is the feature dimension of keys. 
As sequence length increases, attention computation becomes increasingly dominant in the overall computational cost, presenting significant challenges for long-context processing.

\textbf{Arithmetic Intensity}
is the ratio of compute operations to memory accesses. It intrinsically shapes algorithm optimization on hardware. Each GPU has a critical arithmetic intensity determined by its peak compute capability and memory bandwidth, calculated as the ratio of these two hardware limits. For computation tasks, arithmetic intensity above this critical threshold becomes compute-bound~(limited by GPU FLOPS), while below it becomes memory-bound~(limited by memory bandwidth). 

Specifically for causal self-attention mechanism, during training and prefilling phases, batched matrix multiplications and attention computations exhibit high arithmetic intensity, making these stages compute-bound on modern accelerators. In contrast, auto-regressive decoding becomes memory-bandwidth constrained
because it generates one token per forward pass while requiring loading the entire key-value cache,
resulting in low arithmetic intensity. 
This leads to different optimization goals --- reducing computation cost during training and prefilling, while reducing memory access during decoding.

\subsection{Overall Framework}

To leverage the potential of attention with natural sparse pattern, we propose replacing the original key-value pairs \( \mathbf{k}_{:t}, \mathbf{v}_{:t} \) in \cref{equ:attn} with a more compact and information-dense set of representation key-value pairs $\tilde{K}_t, \tilde{V}_t$ given each query \( \mathbf{q}_t \).
Specifically, we formally define the optimized attention output as follows:
\begin{equation}
\tilde{K}_t = f_K(\mathbf{q}_t, \mathbf{k}_{:t}, \mathbf{v}_{:t}), \quad \tilde{V}_t = f_V(\mathbf{q}_t, \mathbf{k}_{:t}, \mathbf{v}_{:t})
\end{equation}
\begin{equation}
\mathbf{o}^*_t=\operatorname{Attn}\left(\mathbf{q}_t,\tilde{K}_t, \tilde{V}_t \right)
\end{equation}

 where 
 \( \tilde{K}_t, \tilde{V}_t \) are dynamically constructed based on the current query \( \mathbf{q}_t \) and the contextual memory \( \mathbf{k}_{:t}, \mathbf{v}_{:t} \). 
We can design various mapping strategies to get different categories of $\tilde{K}_t^c, \tilde{V}_t^c$, and combine them as follows:

\begin{equation}
\mathbf{o}^*_t = \sum_{c \in \mathcal{C}} g_t^c \cdot \text{Attn}(\mathbf{q}_t, \tilde{K}_t^c, \tilde{V}_t^c).
\label{equ:gate_merge}
\end{equation}
As illustrated in \cref{fig:framework}, \method{} have three mapping strategies $\mathcal{C}=\{\text{cmp},\text{slc},\text{win}\}$, representing compression, selection, and sliding window for keys and values.
$g_t^c\in [0, 1]$ is the gate score for corresponding strategy $c$, derived from input features via an MLP and sigmoid activation.
Let \( N_t \) denote the total number of remapped keys/values:
\begin{equation}
N_t = \sum_{c \in \mathcal{C}}\text{size}[\tilde{K}^c_t].
\end{equation}
We maintain a high sparsity ratio by ensuring${N_t}{} \ll t$.

\subsection{Algorithm Design}
In this subsection, we introduce the design of our remapping strategies \( f_K \) and \( f_V \): token compression, token selection, and sliding window.

\subsubsection{Token Compression} 
By aggregating sequential blocks of keys or values into block-level representations, we obtain compressed keys and values that capture the information of the entire block.
Formally, the compressed key representation is defined as:
\begin{equation}
\tilde{K}^\text{cmp}_t = f_K^\text{cmp}(\mathbf{k}_{:t}) = \left\{\phi(\mathbf{k}_{i d+1: i d+l})\middle| 0\leq i\leq\left\lfloor\frac{t-l}{d}\right\rfloor\right\}
\end{equation}
where \( l \) is the block length, \( d \) is the sliding stride between adjacent blocks, and
\( \phi \) is a learnable MLP with intra-block position encoding to map keys in a block to a single compressed key.
$\tilde{K}_t^\text{cmp}\in \mathbb{R}^{ d_k \times \left\lfloor\frac{t-l}{d}\right\rfloor }$ is tensor composed by compresion keys. 
Usually, we adopt $d < l$ to mitigate information fragmentation.
An analogous formulation holds for the compressed value representation \( \tilde{V}_t^\text{cmp} \). 
Compressed representations capture coarser-grained higher-level semantic information and reduce computational burden of attention.

\subsubsection{Token Selection} 
Using only compressed keys, values might lose important fine-grained information, motivating us to selectively preserve individual keys, values. Below we describe our efficient token selection mechanism that identifies and preserves the most relevant tokens with low computational overhead.

\textbf{Blockwise Selection.} 
Our selection strategy processes key and value sequences in spacial continuous blocks, motivated by two key factors: hardware efficiency considerations and inherent distribution patterns of attention scores. \textit{\textbf{Blockwise selection is crucial to achieve efficient computation on modern GPUs.}} That is because modern GPU architectures exhibit significantly higher throughput for continuous block accesses compared to random index-based reads. Also, blockwise computation enables optimal utilization of Tensor Cores. This architectural characteristic has established blockwise memory access and computation as a fundamental principle in high-performance attention implementations, as exemplified by FlashAttention's block-based design. \textit{\textbf{Blockwise selection follows the inherent distribution patterns of attention scores.}} Prior works~\citep{minference} have shown that attention scores often exhibit spatial continuity, suggesting that neighboring keys tend to share similar importance levels. Our visualization in \cref{appendix:vis} also shows this spatial continuous pattern.

To implement blockwise selection, we first divide key, value sequences into selection blocks. To identify the most important blocks for attention computation, we need to assign importance scores to each block. Below we present our method for computing these block-level importance scores.

\textbf{Importance Score Computation.} 
Computing block importance scores could introduce significant overhead. Fortunately, the attention computation of compression tokens produces intermediate attention scores that we can leverage to induce selection block importance scores, formulated as: % can be formulated as:
\begin{equation}
\mathbf{p}_t^\text{cmp} = \operatorname{Softmax}\left(\mathbf{q}_t^T \tilde{K}_t^\text{cmp}\right),
\end{equation}
where \( \mathbf{p}_t^\text{cmp} \in \mathbb{R}^{\left\lfloor\frac{t-l}{d}\right\rfloor+1}  \) is the attention scores between $q_t$ and  compression keys $\tilde{K}_t^\text{cmp}$. 
Let $l'$ denote the selection block size. 
When compression blocks and selection blocks share the same blocking scheme, i.e., $l'=l=d$,
we can directly obtain the selection block importance scores $\mathbf{p}_t^\text{slc}$ by $\mathbf{p}_t^\text{slc} = \mathbf{p}_t^\text{cmp}$ straightforwardly.
For cases where the blocking schemes differ, we derive the importance scores for selection blocks according to their spatial relationship. Given $l\leq l', $ $d \mid l$ and $d \mid l'$, we have:
\begin{equation}
\mathbf{p}_t^\text{slc}[j] = \sum_{m=0}^{\frac{l'}{d}-1}\sum_{n=0}^{\frac{l}{d} -1} \mathbf{p}_t^\text{cmp}\left[\frac{l'}{d}j -m -n \right],
\label{equ.8}
\end{equation}
where$[\cdot]$ denotes the indexing operator for accessing vector element. 
For models employing GQA or MQA where key-value caches are shared across query heads, consistent block selection across these heads has to be ensured to minimize KV cache loading during decoding. The shared importance scores across heads in a group are formally defined as:

\begin{equation}
{\mathbf{p}_t^{\text{slc}}}' = \sum_{h=1}^{H} \mathbf{p}_{t}^{\text{slc}, (h)},
\end{equation}
where $(h)$ in the superscript denotes the head index, and $H$ is the number of query heads in each group. This aggregation ensures consistent block selection across heads within the same group.

\textbf{Top-$\pmb{n}$ Block Selection.} After obtaining the selection block importance scores,
We retain tokens within the top-$n$ sparse blocks ranked by block importance scores, formulated as:
\begin{equation}
\mathcal{I}_t = \{i \mid \text{rank}({\mathbf{p}_t^\text{slc}}'[i]) \leq n\}
\end{equation}
\begin{equation}
\tilde{K}^\text{slc}_t =  \operatorname{Cat}\left[\{\mathbf{k}_{il'+1:(i+1)l'}|i \in \mathcal{I}_t\}\right],
\end{equation}
where rank$(\cdot)$ denotes the ranking position in descending order, with rank = 1 corresponding to the highest score, \( \mathcal{I}_t \) is the set of selected blocks' indices,
\( \operatorname{Cat} \) denotes the concatenation operation.
$\tilde{K}_t^\text{slc}\in \mathbb{R}^{ d_k \times nl' }$ is tensor composed by compresion keys. 
An analogous formulation applies to the fine-grained value \( \tilde{V}^\text{slc}_t \).
The selected keys and values then participate in the attention computation with $\mathbf{q}_t$ as defined in \cref{equ:gate_merge}.

\subsubsection{Sliding Window}
In attention mechanisms, local patterns typically adapt faster and can dominate the learning process, potentially preventing the model from effectively learning from compression and selection tokens. To address this issue, we introduce a dedicated sliding window branch that explicitly handles local context, allowing other branches (compression and selection) to focus on learning their respective features without being shortcutted by local patterns. Specifically, we maintain recent tokens $\tilde{K}_t^\text{win}=\mathbf{k}_{t-w:t}, \tilde{V}_t^\text{win}=\mathbf{v}_{t-w:t}$ in a window $w$, and isolate attention computations of different information sources (compression tokens, and selected tokens, sliding window) into separate branches. These branch outputs are then aggregated through a learned gating mechanism. 
To further prevent shortcut learning across attention branches with marginal computational overhead, we provide independent keys and values for three branches.
This architectural design enables stable learning by preventing gradient interference between local and long-range pattern recognition, while introducing minimal overhead.

After obtaining all three categories of keys and values ($\tilde{K}_t^\text{cmp}, \tilde{V}_t^\text{cmp}$; $\tilde{K}_t^\text{slc}, \tilde{V}_t^\text{slc}$; and $\tilde{K}_t^\text{win}, \tilde{V}_t^\text{win}$), we compute the final attention output following \cref{equ:gate_merge}.
Together with the compression, selection, and sliding window mechanisms described above, this forms the complete algorithmic framework of \method{}.

\subsection{Kernel Design}
\begin{figure}[h] 
\centering 
\includegraphics[width=0.5\linewidth]{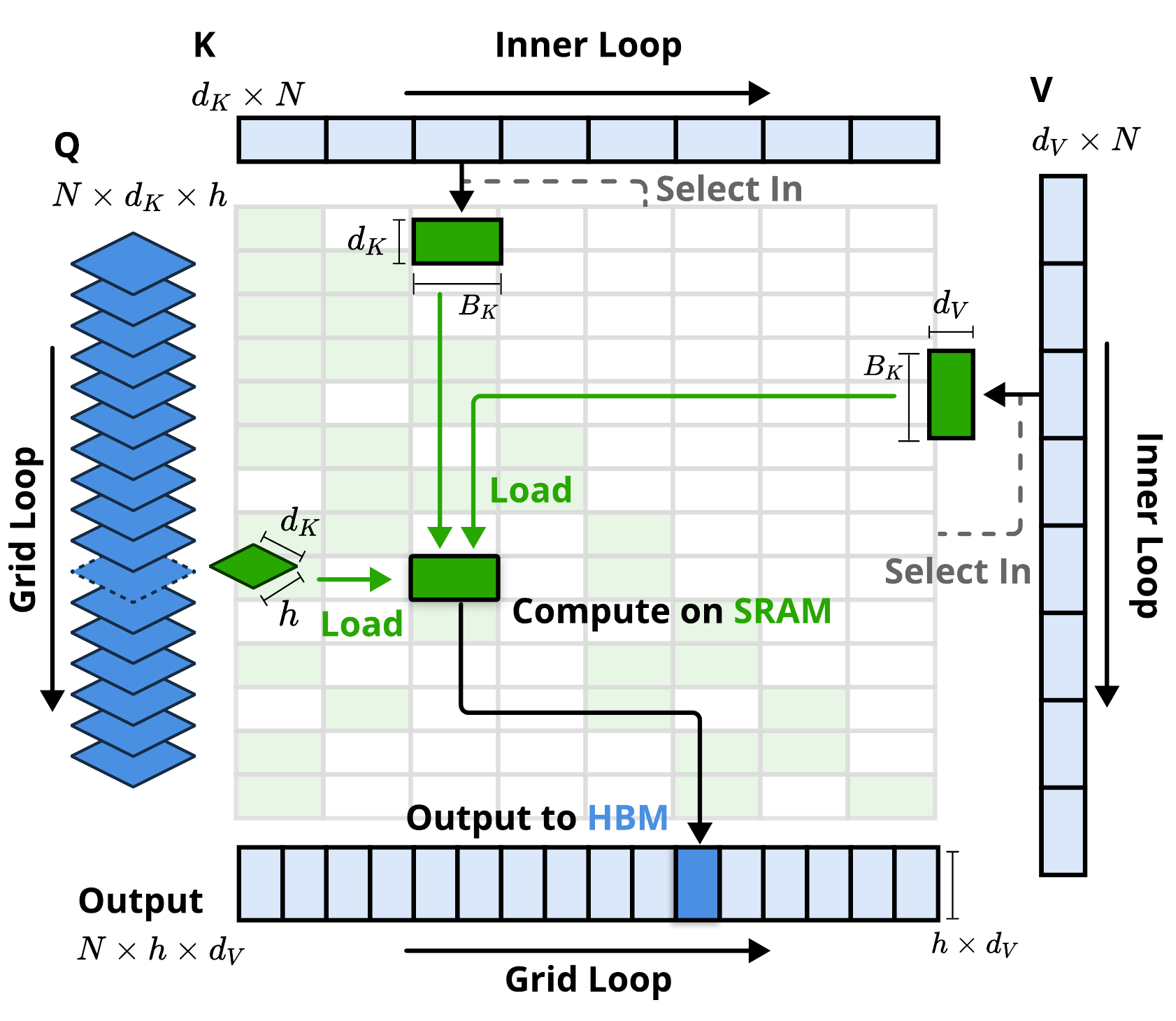}
\caption{Kernel design for \method{}. The kernel loads queries by GQA groups (Grid Loop), fetches corresponding sparse KV blocks (Inner Loop), and performs attention computation on SRAM. Green blocks indicate data on SRAM, while blue indicates data on HBM.}
\label{fig:fig3} 
\end{figure}

To achieve FlashAttention-level speedup during the training and prefilling, we implement hardware-aligned sparse attention kernels upon Triton. Given MHA is memory-intensive and inefficient for decoding, we focus on architectures with shared KV caches like GQA and MQA following the current state-of-the-art LLMs. While compression and sliding window attention computations are readily compatible with existing FlashAttention-2 kernels, we introduce the specialized kernel design for sparse selection attention.
If we were to follow FlashAttention's strategy of loading temporally continuous query blocks into SRAM, it would result in inefficient memory access since queries within a block may require disjoint KV blocks. To address this, our key optimization lies in a different query grouping strategy: for each position on the query sequence, we load all query heads within a GQA group (they share the same sparse KV blocks) into SRAM. \cref{fig:fig3} illustrates our forward pass implementation.
The proposed kernel architecture is characterized by the following key features:
\begin{enumerate}
    \item \textbf{Group-Centric Data Loading}. For each inner loop, load all heads' queries $Q\in \mathbb{R}^{[ h, d_k]}$ in the group at position $t$ and their shared sparse key/value block indices $\mathcal{I}_t$.
    \item \textbf{Shared KV Fetching}. In the inner loop, Sequentially load continuous key/value blocks indexed by $\mathcal{I}_t$ into SRAM as $K\in\mathbb{R}^{[B_k, d_k]}, V\in\mathbb{R}^{[B_k, d_v]}$ to minimize memory loading, where $B_k$ is the kernel block size satisfying $B_k | l'$.
    \item \textbf{Outer Loop on Grid}. Since the inner-loop length (proportional to the selected block count $n$) remains nearly identical for different query blocks, we put query/output loops in Triton's grid scheduler to simplify and optimize the kernel.
\end{enumerate}

This design achieves near-optimal arithmetic intensity by (1) eliminating redundant KV transfers through group-wise sharing, and (2) balancing compute workloads across GPU streaming multiprocessors.

\section{Experiments}
\label{sec:experiment}
We evaluate \method{} through three lenses: (1) general benchmarks performance, (2) long-context benchmarks performance, and (3) chain-of-thought reasoning performance, comparing against Full Attention baseline and state-of-the-art sparse attention methods. We defer the efficiency analysis of our sparse computation paradigm to \cref{sec:time_comparsion}, where we provide detailed discussions on training and inference speed.

\subsection{Pretraining Setup} 
\begin{figure}[t]
    \centering
    \includegraphics[width=0.8\linewidth]{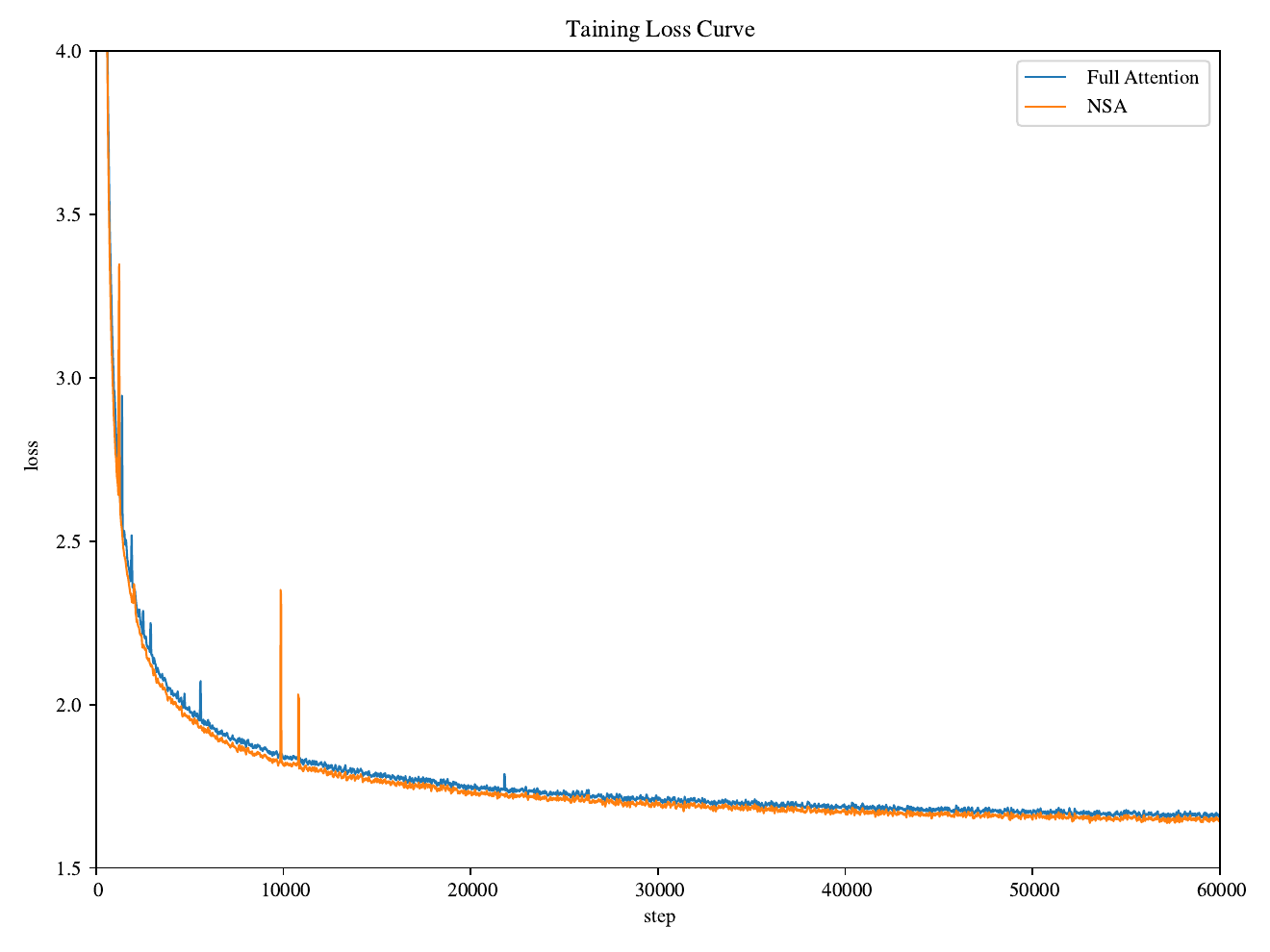}
    \caption{Pretraining loss comparison between Full Attention and our \method{} on 27B-parameter model. Both models exhibit stable convergence, with \method{} achieving lower loss values.}
    
    \label{fig:losscurve}
\end{figure}

Following the common practice in state-of-the-art LLMs, our experiments adopt a backbone combining Grouped-Query Attention (GQA) and Mixture-of-Experts (MoE), featuring $27\text{B}$ total parameters with $3\text{B}$ active parameters. The model consists of 30 layers with a hidden dimension of 2560. For GQA, we set the number of groups to 4, with a total of 64 attention heads. For each head, the hidden dimensions of the query, key, and value are configured as $d_q = d_k = 192$ and $d_v = 128$, respectively. For MoE, we utilize the DeepSeekMoE~\citep{dai2024deepseekmoe,deepseekV2} structure, with 72 routed experts and 2 shared experts, and set the top-k experts to 6. To ensure training stability, the MoE in the first layer is replaced by an MLP in the form of SwiGLU. The proposed architecture achieves an effective trade-off between computation cost and model performance.
For \method{}, we set compression block size $l=32$, sliding stride $d=16$, selected block size $l'=64$, selected block count $n=16$ (including fixed activating the 1 initial block and 2 local blocks), and sliding window size $w=512$. Both Full Attention and sparse attention models are pretrained on $270\text{B}$ tokens of $8\text{k}$-length texts, followed by continued training and supervised fine-tuning on $32\text{k}$-length texts with YaRN~\citep{yarn} to achieve long-context adaptation. Both models are trained to full convergence to ensure fair comparison. As shown in ~\cref{fig:losscurve}, the pretraining loss curve of our \method{} and Full Attention baseline demonstrates stable and smooth decline, with \method{} consistently outperforming the Full Attention model.
\begin{table*}[t] 
\centering 
\tabcolsep=4pt
\resizebox{\textwidth}{!}{
\begin{tabular}{lcccccccccl} 
\toprule 
\multirow{2}*{Model} 
& \multicolumn{1}{c}{MMLU} & \multicolumn{1}{c}{MMLU-PRO} & \multicolumn{1}{c}{CMMLU} & \multicolumn{1}{c}{BBH} & \multicolumn{1}{c}{GSM8K} & \multicolumn{1}{c}{MATH} & \multicolumn{1}{c}{DROP} & \multicolumn{1}{c}{MBPP} & \multicolumn{1}{c}{HumanEval} 
& \multirow{2}*{Avg.}\\ 
& \scriptsize{Acc. 5-shot} & \scriptsize{Acc. 5-shot} & \scriptsize{Acc. 5-shot} & \scriptsize{Acc. 3-shot} & \scriptsize{Acc. 8-shot} & \scriptsize{Acc. 4-shot} & \scriptsize{F1 1-shot} & \scriptsize{Pass@1 3-shot} & \scriptsize{Pass@1 0-shot} & \\
\midrule 
Full Attn & \textbf{0.567} & 0.279 & 0.576 & 0.497 & 0.486 & 0.263 & 0.503 & \textbf{0.482} & 0.335  &0.443
\\ 
\method{} & 0.565 & \textbf{0.286} & \textbf{0.587} & \textbf{0.521} & \textbf{0.520} & \textbf{0.264} & \textbf{0.545} & 0.466 & \textbf{0.348}  &\textbf{0.456}\\ 
\bottomrule 
\end{tabular} 
}
\caption{Pretraining performance comparison between the full attention baseline and \method{} on general benchmarks, across knowledge (MMLU, MMLU-PRO, CMMLU), reasoning (BBH, GSM8K, MATH, DROP), and coding (MBPP, HumanEval) tasks. \method{} achieves superior average performance  on most benchmarks despite high sparsity.}
\label{tab:base_eval}
\end{table*}

\subsection{Baselines Methods} 
In addition to comparing with Full Attention, we evaluate several state-of-the-art inference-stage sparse attention methods: H2O \citep{h2o}, infLLM \citep{infllm}, Quest \citep{quest}, and Exact-Top, which first computes full attention score and select the top-$n$ scores keys corresponding to each query and then calculates attention on these positions. These methods span diverse sparse attention paradigms, including KV-cache eviction, query-aware selection, and exact top-$n$ sparse selection. 

For general evaluation, where most samples have lengths within the local context window of sparse attention baselines, these methods are effectively equivalent to Full Attention. Therefore, we present only the comparison results between \method{} and Full Attention baseline in this setting. In the long-context evaluation, we conduct comparisons across all baseline methods, with the sparsity of all sparse attention methods set to the same to ensure a fair comparison. For chain-of-thought reasoning evaluation, which requires long-text supervised fine-tuning, we limit our comparison to Full Attention, as sparse attention baselines do not support training.

\subsection{Performance Comparison}
\begin{table*}[t]
\centering
\tabcolsep=4pt
\resizebox{\textwidth}{!}{
\begin{tabular}{lccccccccccc}
\toprule
\multirow{2}*{Model} & \multicolumn{3}{c}{{SQA}} & \multicolumn{4}{c}{{MQA}} & \multicolumn{2}{c}{{Synthetic}} & \multicolumn{1}{c}{{Code}} & \multirow{2}*{Avg.}\\
\cmidrule(lr){2-4} \cmidrule(lr){5-8} \cmidrule(lr){9-10} \cmidrule(lr){11-11}
 & MFQA-en & MFQA-zh & Qasper & HPQ & 2Wiki & GovRpt & Dur & PassR-en & PassR-zh & LCC & \\
\midrule
H2O & 0.428& 0.429& 0.308& 0.112& 0.101& 0.231& 0.208& 0.704& 0.421& 0.092& 0.303
\\
InfLLM & 0.474& 0.517& 0.356& 0.306& 0.250& 0.277& 0.257& 0.766& 0.486& 0.143& 0.383
\\
Quest & 0.495& 0.561& 0.365& 0.295& 0.245& 0.293& 0.257& 0.792& 0.478& 0.135& 0.392
\\
Exact-Top& 0.502& 0.605& 0.397& 0.321& 0.288& \underline{0.316}& 0.291& 0.810& 0.548& 0.156& 0.423\\
Full Attn & \textbf{0.512} & \underline{0.623}& \underline{0.409}& \underline{0.350} & \underline{0.305}& \textbf{0.324} & \underline{0.294}& \underline{0.830}& \textbf{0.560} & \underline{0.163} & \underline{0.437}\\
\midrule
\method{} & \underline{0.503}& \textbf{0.624} & \textbf{0.432} & \textbf{0.437} & \textbf{0.356}& {0.307}& \textbf{0.341} & \textbf{0.905}& \underline{0.550}& \textbf{0.232}& \textbf{0.469}\\
\bottomrule
\end{tabular}
}
\caption{Performance comparison between our \method{} and baselines on LongBench, including subsets in single document QA, multi-document QA, synthetic and code task categories. \method{} outperformed most of the baselines including Full Attention.}
\label{tab:longbench}

\end{table*}

\textbf{General Evaluation.} We evaluated the pretrained \method{} and Full Attention baseline, on a comprehensive suite of benchmarks spanning knowledge, reasoning, and coding capabilities, including MMLU~\citep{hendrycks2020measuring}, MMLU-PRO~\citep{wang2024mmlu}, CMMLU~\citep{li2023cmmlu}, BBH~\citep{suzgun2022challenging}, GSM8K~\citep{cobbe2021training}, MATH~\citep{hendrycks2020measuring}, DROP~\citep{dua2019drop}, MBPP~\citep{austin2021program}, and HumanEval~\citep{chen2021evaluating}. The results are shown in \cref{tab:base_eval}. Despite its sparsity, \method{} achieves superior overall performance, outperforming all baselines including Full Attention on 7 out of 9 metrics. This indicates that although \method{} may not fully leverage its efficiency advantages on shorter sequences, it shows strong performance.
Notably, \method{} demonstrates significant gains in reasoning-related benchmarks (DROP: +0.042, GSM8K: +0.034), suggesting that 
our pretraining helps models to develop specialized attention mechanisms. This sparse attention pretraining mechanism forces model to focus on the most important information, potentially enhancing performance by filtering out noise from irrelevant attention pathways.
The consistent performance across diverse evaluations also validates \method{}'s robustness as a general-purpose architecture.

\begin{figure}[t]
    \centering
    \includegraphics[width=0.7\textwidth]{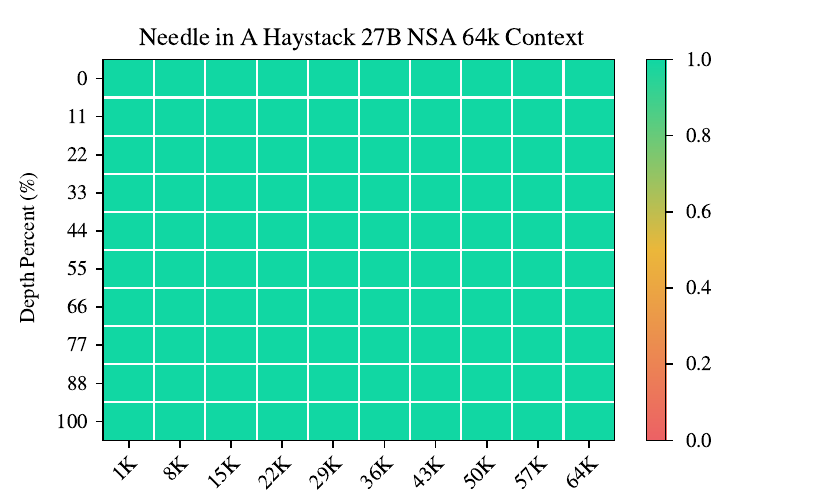}
    \caption{Needle-in-a-Haystack retrieval accuracy across context positions with 64k context length. \method{} achieves perfect accuracy through its hierarchical sparse attention design.}
    \label{fig:niah}
\end{figure}

\textbf{Long-Context Evaluation.} As shown in \cref{fig:niah},
\method{} achieves perfect retrieval accuracy across all positions in 64k-context needle-in-a-haystack~\citep{kamradt2023llmtest} test. This performance stems from our hierarchical sparse attention design, which combines compression tokens for efficient global context scanning, and  
selection tokens for precise local information retrieval.  
The coarse-grained compression identifies relevant context blocks at low computational cost, while the token-level attention on selected tokens ensures the preservation of critical fine-grained information. This design enables \method{} to maintain both global awareness and local precision.

We also evaluate \method{} on LongBench~\citep{bai2023longbench} against state-of-the-art sparse attention methods and Full Attention baseline. 
To ensure consistent sparsity, we set the token activated by each query in all sparse attention baselines to 2560 tokens, which corresponds to the average number of tokens activated in \method{} when handling 32k sequence lengths. Following StreamLLM~\citep{streaming}, this token budget includes the leading 128 tokens and 512 local tokens.
We exclude certain subsets from LongBench due to their low scores across all models, which may not provide meaningful comparisons. As shown in \cref{tab:longbench}, \method{} achieves the highest average score 0.469, outperforming all baselines (+0.032 over Full Attention and +0.046 over Exact-Top). This improvement arises from two key innovations: (1) our native sparse attention design, which enables end-to-end optimization of sparse patterns during pretraining, facilitates synchronized adaptation between the sparse attention module and other model components; and (2) the hierarchical sparse attention mechanism achieves a balance between local and global information processing.

Notably, \method{} demonstrates exceptional performance on tasks requiring complex reasoning over long contexts, achieving +0.087 and +0.051 improvements over Full Attention on multi-hop QA tasks (HPQ and 2Wiki), exceeding the performance of baselines on code understanding (LCC: +0.069), and outperforming other methods on passage retrieval (PassR-en: +0.075). These results validate \method{}'s capability to handle diverse long-context challenges, with its natively pretrained sparse attention providing additional benefits in learning task-optimal patterns.

\textbf{Chain-of-Thought Reasoning Evaluation.}  
To evaluate \method{}'s compatibility with advanced downstream training paradigms, we investigate its capacity to acquire chain-of-thought mathematical reasoning abilities via post-training. 
Given the limited effectiveness of reinforcement learning on smaller-scale models, we employ knowledge distillation from DeepSeek-R1, conducting supervised fine-tuning (SFT) with 10B tokens of 32k-length mathematical reasoning traces. This produces two comparable models: Full Attention-R (Full Attention baseline) and \method{}-R (our sparse variant). We assess both models on the challenging American Invitational Mathematics Examination (AIME 24) benchmark. We use a sampling temperature of 0.7 and a top-$p$ value of 0.95 to generate 16 responses for each question and obtain the average score. To validate the impact of reasoning depth, we conduct experiments with two generation context limits: 8k and 16k tokens, measuring whether extended reasoning chains improve accuracy. Example comparisons of model predictions are provided in \cref{app:aime}.
\begin{table}[t]
\centering
\resizebox{0.55\textwidth}{!}{
\begin{tabular}{lcc}
\toprule
Generation Token Limit ~~~ & ~~ 8192 ~~  & ~~ 16384 ~~  \\
\midrule
Full Attention-R & 0.046 & 0.092 \\
\method{}-R & \textbf{0.121} & \textbf{0.146} \\
\bottomrule
\end{tabular}
}
\caption{AIME Instruction-based Evaluating after supervised fine-tuning. Our \method{}-R demonstrates better performance than Full Attention-R at both 8k and 16k sequence lengths}

\label{tab:aime}
\end{table}

As shown in \cref{tab:aime}, \method{}-R achieves significantly higher accuracy than Full Attention-R under the 8k context setting (+0.075), with this advantage persisting at 16k contexts (+0.054).
These results validate two key benefits of native sparse attention: (1) The pretrained sparse attention patterns enable efficient capture of long-range logical dependencies critical for complex mathematical derivations; (2) Our architecture's hardware-aligned design maintains sufficient context density to support growing reasoning depth without catastrophic forgetting. The consistent outperformance across context lengths confirms sparse attention's viability for advanced reasoning tasks when natively integrated into the training pipeline.

\section{Efficiency Analysis }
\label{sec:time_comparsion}
We evaluate the computational efficiency of \method{} against Full Attention on an 8-GPU A100 system. In efficiency analysis, we also configure the model with GQA group $g=4$, heads per group $h=16$, query/key dimension $d_k=192$, and value dimension $d_v=128$. 
Following the same settings in \cref{sec:experiment}, we set \method{} compression block size $l=32$, sliding stride $d=16$, selected block size $l'=64$, selected block count $n=16$, and sliding window size $w=512$.

\begin{figure}[t]
    \centering
    \includegraphics[width=0.65\textwidth]{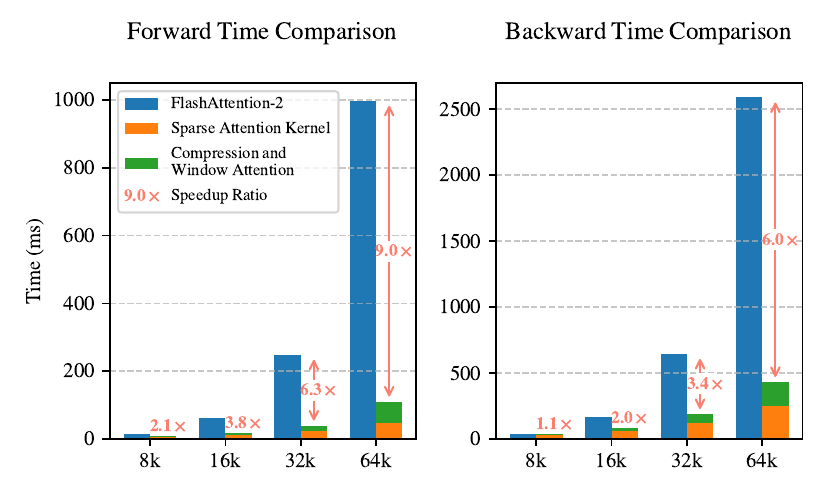}
    \caption{Comparison of Triton-based \method{} kernel with Triton-based FlashAttention-2 kernel. Our implementation significantly reduces latency across all context lengths, with the improvement becoming more pronounced as input length increases.}
    \label{fig:speed}
\end{figure}
\subsection{Training Speed}
We compare the Triton-based implementations of our \method{} attention and Full Attention with Triton-based FlashAttention-2 to ensure fair speed comparison across the same backend.
As shown in \cref{fig:speed}, our \method{} achieves progressively greater speedups as context length increases, up to 9.0$\times$ forward and 6.0$\times$ backward speedup at 64k context-length. Notably, the speed advantage becomes more pronounced with longer sequences. This speedup stems from our hardware-aligned algorithm design to maximize the efficiency of sparse attention architecture: (1) The Blockwise memory access pattern maximizes Tensor Core utilization through coalesced loads, (2) The 
delicate loop scheduling in the kernel eliminates redundant KV transfers.

\begin{table}[t]
\centering
\resizebox{0.66\textwidth}{!}{
\begin{tabular}{lcccc}
\toprule
Context Length& ~~8192~~& ~~ 16384 ~~ & 32768 &~~65536~~\\
\midrule
 Full Attention& ~~8192~~& 16384& 32768 &~~65536~~\\
\method{}& ~~2048~~&  2560& 3584 &~~5632~~\\
\midrule
Expected Speedup& \textbf{4$\times$}& \textbf{~~6.4$\times$}& \textbf{~~9.1$\times$} &\textbf{~~11.6$\times$} \\
\bottomrule
\end{tabular}
}
\caption{Memory access volume (in equivalent number of tokens) per attention operation during decoding. Due to the low arithmetic intensity and memory-bound nature of decoding, the expected speedup is approximately linear with the volume of memory access.}

\label{tab:infer}
\end{table}

\subsection{Decoding Speed}
The decoding speed of Attention is primarily determined by the memory access bottleneck, which is closely tied to the amount of KV cache loading. In each decoding step, Our \method{} just needs to load at most $\left\lfloor\frac{s-l}{d}\right\rfloor$ compression tokens, $nl'$ selected tokens, and $w$ neighbor tokens, where $s$ is the cached sequence length.
As shown in \cref{tab:infer}, our method exhibits a significant reduction in latency as the decoding length increases, achieving up to 11.6$\times$ speedup at 64k context-length. This advantage in memory access efficiency also amplifies with longer sequences.

\section{Discussion}
In this section, we reflect on the development process of \method{} and discuss key insights gained from our exploration of different sparse attention strategies. While our approach demonstrates promising results, understanding the challenges encountered with alternative strategies and analyzing attention patterns provides valuable context for future research directions. We first examine challenges with alternative token selection strategies that motivated our design choices, followed by visualizations that offer insights into attention distribution patterns.

\subsection{Challenges with Alternative Token Selection Strategies}
\label{app:challenge}
Before designing \method{}, we explored adapting existing sparse attention methods to the training stage. However, these attempts encountered various challenges, prompting us to design a different sparse attention architecture:

\textbf{Key-Clustering Based Strategies.} 
We examined clustering-based strategies like ClusterKV~\citep{clusterkv}. These methods store Keys and Values from the same cluster in contiguous memory regions. While theoretically feasible for training and inference, they face three significant challenges: (1) Non-trivial computational overhead introduced by dynamic clustering mechanisms; (2) Operator optimization difficulties exacerbated by inter-cluster imbalances, especially in Mixture-of-Experts (MoE) systems, where skewed Expert Parallelism (EP) group execution times lead to persistent load imbalances; (3) Implementation constraints arising from the need for mandatory periodic reclustering and chunk-sequential training protocols. These combined factors create substantial bottlenecks, significantly limiting their effectiveness for real-world deployment.
\begin{figure}[t]
    \centering
    \begin{minipage}{0.52\textwidth}
        \centering
        \includegraphics[width=\linewidth]{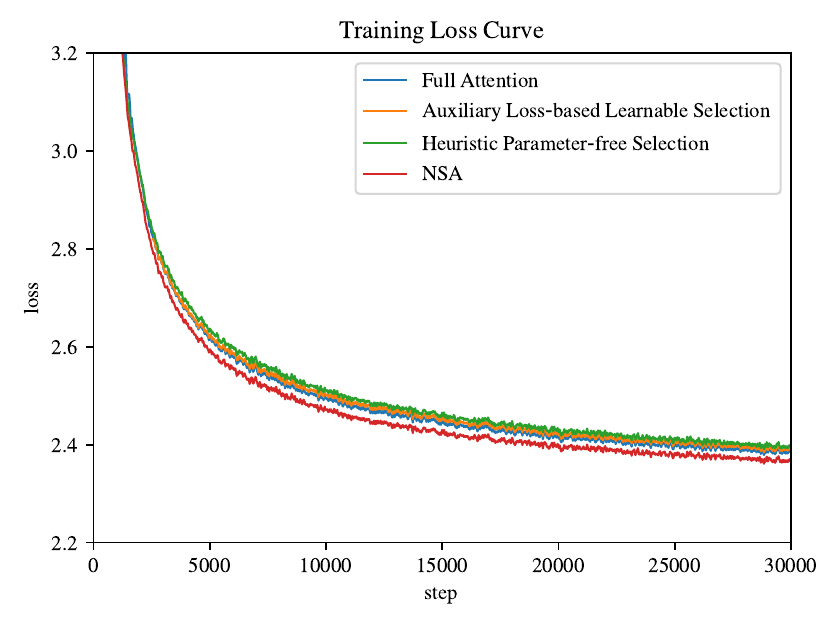}
        \caption{Compare training loss on a 3B-parameter model with Full Attention and different token selection strategies and. Our \method{} achieves better performance.}
        \label{fig:losscurve3b}
    \end{minipage}
    \hfill
    \begin{minipage}{0.44\textwidth}
        \centering
        \includegraphics[width=\linewidth]{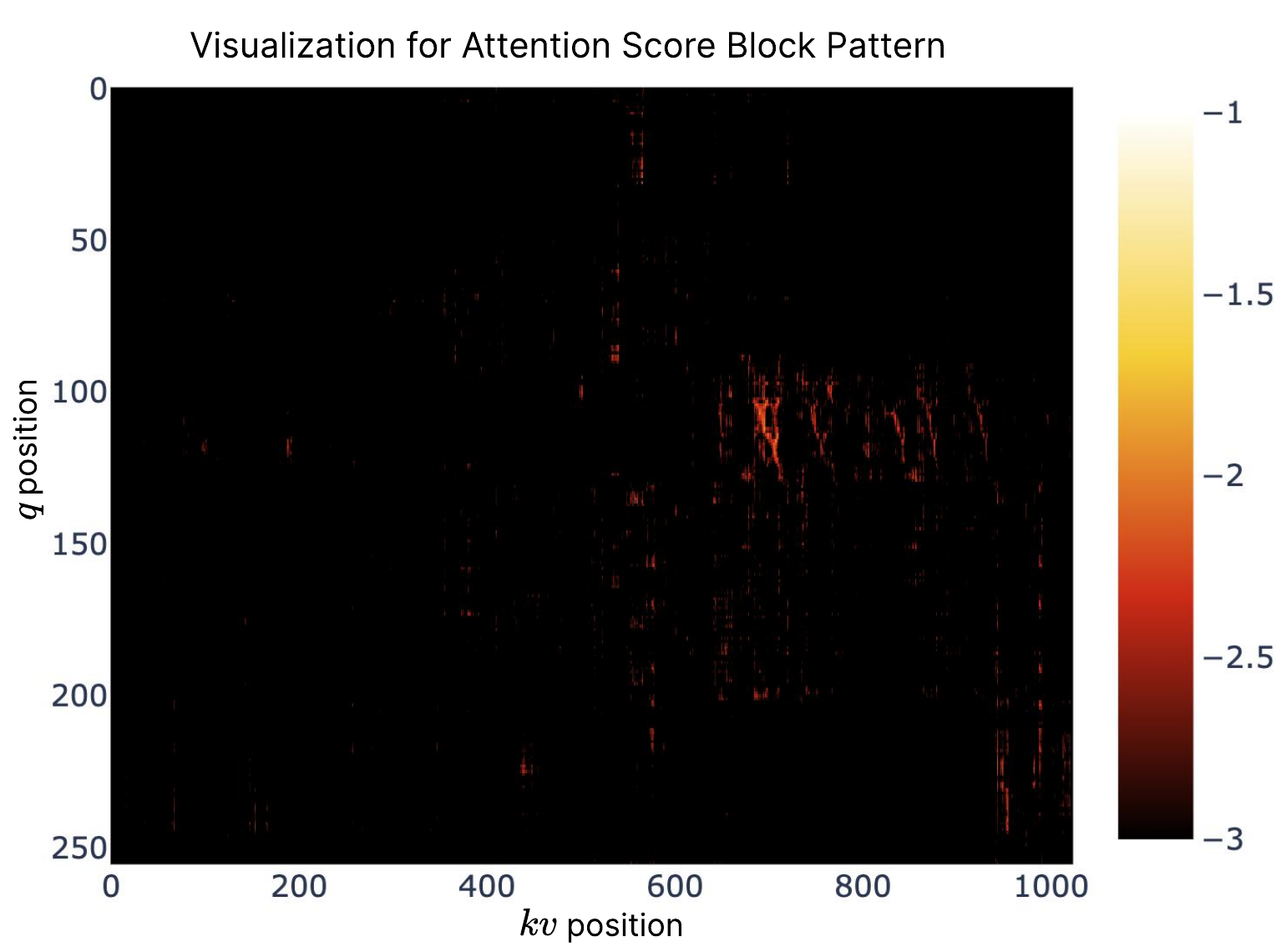}
        \caption{Visualization of Attention Map on a Full Attention transformer. Light-colored regions indicate higher attention values. As shown in the figure, attention scores exhibit blockwise clustering distribution.}
        \label{fig:visual}
    \end{minipage}
\end{figure}

\textbf{Other Blockwise Selection Strategies.}
We also considered blockwise key, value selection strategies different from \method{}, such as Quest~\citep{quest} and InfLLM~\citep{infllm}. These methods rely on
computing an importance score
for each KV block and selecting the top-$n$ blocks based on their similarity with $q_t$. However, existing methods face two critical issues: (1) Since the selection operation is non-differentiable, importance score computation based on neural networks relies on auxiliary loss, which increases operator overhead and often degrades model performance; (2) Heuristic parameter-free importance score computation strategy suffer from low recall rates, leading to suboptimal performance. We evaluate both approaches on a 3B-parameter model with similar architecture and compare their loss curve with \method{} and Full Attention.
For the auxiliary loss-based selection method, we introduce additional queries for each token and representative keys for each block to estimate the block importance scores.
We compute block-level supervision signals by mean-pooling attention scores within each key block, and use KL divergence to supervise block importance prediction. We maintain individual query granularity instead of block-averaged queries to accommodate efficient decoding. This auxiliary loss-based importance estimation shares conceptual similarity with SeerAttention~\citep{gao2024seerattention}.
For the heuristic parameter-free selection method, following the strategy of Quest, we implement direct selection using the product between queries and coordinate-wise min-max of the key chunks, without introducing additional parameters. We also explore a cold-start training approach where Full Attention is applied for the initial 1000 steps before transitioning to the heuristic blockwise selection.
As shown in ~\cref{fig:losscurve3b}, both methods exhibited inferior loss.

\vspace{-10pt}
\subsection{Visualization}
\label{appendix:vis}

To explore potential patterns in transformer attention distributions and seek inspiration for our design, we visualize the attention map from our pretrained 27B Full Attention model in \cref{fig:visual}. The visualization reveals interesting patterns where attention scores tend to exhibit blockwise clustering characteristics, with nearby keys often showing similar attention scores. This observation inspired our design of \method{}, suggesting that selecting key blocks based on spatial continuity might be a promising approach. The blockwise clustering phenomenon indicates that tokens adjacent in the sequence may share certain semantic relationships with query tokens, though the exact nature of these relationships requires further investigation. This observation motivated us to explore a sparse attention mechanism that operates on continuous token blocks rather than individual tokens, aiming to enhance computational efficiency and preserve high-attention patterns. 
\section{Related Works}
\label{app:related}

We review existing approaches that improve the efficiency of attention computation through sparse attention. These methods can be broadly categorized into three groups based on their core strategies: (1) fixed sparse pattern, (2) dynamic token pruning, and (3) query-aware selection. We introduce several representative works from each category.

\subsection{Fixed Sparse Pattern}
SlidingWindow is a commonly used approach that allows the query to compute attention only within a fixed window. 
StreamingLLM~\citep{streaming} combines attention sinks with local windows to process continuous text streams. 
MoA~\citep{moa} and DuoAttention~\citep{xiao2024duoattention} also adopt similar local and sink information for long sequence modeling.
Longformer~\citep{beltagy2020longformer} interleaves local windowed attention with global tokens to process long sequences.
Compared with them, our \method{} does not rely on pre-defined sparse patterns, but learns the patterns automatically, unlocking the potential to utilize full context. 

\subsection{Dynamic Token Pruning}
Some methods are designed to reduce memory and computation costs during inference through dynamic KV-cache pruning.
H2O~\citep{h2o}, BUZZ~\citep{zhao2024buzz}, and SepLLM~\citep{chen2024sepllm} implement adaptive approaches to reduce KV-cache memory usage during decoding. These methods dynamically evict tokens deemed less important for future predictions.
FastGen~\cite{ge2023model} and HeadKV~\citep{fu2024not} optimize computation by assigning different strategies to individual attention heads.
SnapKV~\citep{snapkv} introduces a token pruning strategy that reduces the KV-cache by selectively retaining only the most crucial features, enabling efficient memory usage. 
Unlike these inference-focused approaches, our \method{} incorporates sparsity natively during the training phase.

\subsection{Query-Aware Selection}
Other works focus on query-dependent token selection methods to reduce attention computation while preserving attention quality.
Quest~\citep{quest} employs a blockwise selection strategy where each chunk's importance is estimated by product between query and coordinate-wise min-max of the key chunks.
InfLLM~\citep{infllm} combines fixed patterns with retrieval by maintaining attention sinks, local context, and retrievable chunks. This method selects representative keys from each chunk to estimate chunk importance.
HashAttention~\citep{desai2024hashattention} formulates pivotal token identification as a recommendation problem by mapping queries and keys to Hamming space using learned functions.
ClusterKV~\citep{clusterkv} achieves sparsity by firstly clustering keys and then selecting the most relevant clusters for attention computation based on query-cluster similarity.
MInference~\citep{minference} and TokenSelect~\citep{wu2024tokenselect} select KV pairs for computation based on token-level importance scoring.
SeerAttention~\citep{gao2024seerattention} separates queries and keys into spatial blocks and performs blockwise selection for efficient computation.
Compared to these methods, our \method{} achieves hardware-aligned sparse attention computation throughout the full model lifecycle, including training, prefilling, and decoding.
\section{Conclusion}
We present \method{}, a hardware-aligned sparse attention architecture for efficient long-context modeling. By integrating hierarchical token compression with blockwise token selection within a trainable architecture, our architecture achieves accelerated training and inference while maintaining Full Attention performance. \method{} advances the state-of-the-art by demonstrating general benchmark performance matches full-attention baselines, exceeding modeling capability in long-context evaluations, and enhanced reasoning ability, all accompanied by measurable reductions in computational latency and achieving significant speedup. 

\bibliography{main}

\newpage
\appendix
\section{Example of AIME Results}
\label{app:aime}

\newcommand{\testcaseheader}[1]{%
    \noindent\colorbox{gray!10}{\makebox[\linewidth][l]{\textbf{#1}}}%
}

\newcommand{\testcase}[3]{%
    \begin{mdframed}[
        linewidth=1pt,
        linecolor=gray!30,
        backgroundcolor=white,
        roundcorner=4pt,
        innertopmargin=8pt,
        innerbottommargin=8pt
    ]
    \testcaseheader{Prompt:}
    \vspace{0.5em}
    \noindent #1
    
    \vspace{1em}
    \testcaseheader{Our Result:}
    \vspace{0.5em}
    \noindent #2
    
    \vspace{1em}
    \testcaseheader{Baseline Result:}
    \vspace{0.5em}
    \noindent #3
    \end{mdframed}
    \vspace{1em}
}

\testcase{
    ``\textbar User\textbar Let $x,y$ and $z$ be positive real numbers that satisfy the following system of equations:
    \[
    \log_2\left(\frac{x}{yz}\right) = \frac{1}{2}
    \]
    \[
    \log_2\left(\frac{y}{xz}\right) = \frac{1}{3}
    \]
    \[
    \log_2\left(\frac{z}{xy}\right) = \frac{1}{4}
    \]
    Then the value of $\left|\log_2(x^4y^3z^2)\right|$ is $\frac{m}{n}$ where $m$ and $n$ are relatively prime positive integers. Find $m+n$.
    Please reason step by step, and put your final answer within \(\boxed{}\).\textbar Assistant\textbar''
}{%
    \begin{quote}
    $\langle$think$\rangle$...$\langle$/think$\rangle$ (2275 tokens)

    Given the system of equations:

    1. $\log_2\left(\frac{x}{yz}\right) = \frac{1}{2}$
    2. $\log_2\left(\frac{y}{xz}\right) = \frac{1}{3}$
    3. $\log_2\left(\frac{z}{xy}\right) = \frac{1}{4}$

    We convert these logarithmic equations to their exponential forms:

    1. $\frac{x}{yz} = \sqrt{2}$
    2. $\frac{y}{xz} = 2^{1/3}$
    3. $\frac{z}{xy} = 2^{1/4}$

    Expressing each variable in terms of $x$, $y$, and $z$:

    1. $x = \sqrt{2} \cdot yz$
    2. $y = 2^{1/3} \cdot xz$
    3. $z = 2^{1/4} \cdot xy$

    Substituting these expressions into each other, we find:

    - From $x = \sqrt{2} \cdot yz$ and $y = 2^{1/3} \cdot xz$, we get $x = \sqrt{2} \cdot (2^{1/3} \cdot xz) \cdot z$, leading to $z = 2^{-5/12}$.
    - Substituting $z$ into the equation for $y$, we get $y = 2^{-3/8}$.
    - Substituting $z$ and $y$ into the equation for $x$, we get $x = 2^{-7/24}$.

    Next, we compute $x^4 y^3 z^2$:

    - $x^4 = (2^{-7/24})^4 = 2^{-7/6}$
    - $y^3 = (2^{-3/8})^3 = 2^{-9/8}$
    - $z^2 = (2^{-5/12})^2 = 2^{-5/6}$

    Multiplying these together:

    \[x^4 y^3 z^2 = 2^{-7/6} \cdot 2^{-9/8} \cdot 2^{-5/6} = 2^{-75/24} = 2^{-25/8}\]

    Taking the logarithm base 2:

    \[\log_2(x^4 y^3 z^2) = -25/8\]

    The absolute value is:

    \[| \log_2(x^4 y^3 z^2) | = 25/8\]

    Thus, $m = 25$ and $n = 8$, giving $m + n = 33$.

    \[
    \boxed{33}
    \]
    \end{quote}
}{%
    \begin{quote}
    $\langle$think$\rangle$...$\langle$/think$\rangle$ (9392 tokens)

    Given the system of logarithmic equations:

    \[
    \log_2\left(\frac{x}{yz}\right) = \frac{1}{2}
    \]
    \[
    \log_2\left(\frac{y}{xz}\right) = \frac{1}{3}
    \]
    \[
    \log_2\left(\frac{z}{xy}\right) = \frac{1}{4}
    \]

    We convert these equations to exponential form:

    \[
    \sqrt{2} = \frac{x}{yz} \implies x = \sqrt{2} \cdot yz
    \]
    \[
    2^{\frac{1}{3}} = \frac{y}{xz} \implies y = 2^{\frac{1}{3}} \cdot xz
    \]
    \[
    2^{\frac{1}{4}} = \frac{z}{xy} \implies z = 2^{\frac{1}{4}} \cdot xy
    \]

    Substituting these into each other, we find:

    1. $x = \sqrt{2} \cdot yz$
    2. $y = 2^{\frac{1}{3}} \cdot xz$
    3. $z = 2^{\frac{1}{4}} \cdot xy$

    By solving these equations step-by-step, we find:

    - $z = 2^{-5/12}$
    - $xy = 2^{-2/3}$
    - $y = 2^{-1/24}$
    - $x = 2^{1/24}$

    However, upon verifying the third equation, we realize the system might not have a solution. Therefore, we use matrix inversion to solve the system:

    \[
    \begin{cases}
    \log_2(x) - \log_2(y) - \log_2(z) = \frac{1}{2} \\
    \log_2(y) - \log_2(x) - \log_2(z) = \frac{1}{3} \\
    \log_2(z) - \log_2(x) - \log_2(y) = \frac{1}{4}
    \end{cases}
    \]

    Let $A = \log_2(x)$, $B = \log_2(y)$, $C = \log_2(z)$. The system becomes:

    \[
    \begin{cases}
    A - B - C = \frac{1}{2} \\
    B - A - C = \frac{1}{3} \\
    C - A - B = \frac{1}{4}
    \end{cases}
    \]

    We solve this system using Cramer's rule and matrix inversion. The determinant of the matrix is -2, and the solutions are:

    \[
    A = \frac{61}{48}, \quad B = \frac{13}{16}, \quad C = -\frac{1}{6}
    \]

    Thus, we have:

    \[
    x = 2^{61/48}, \quad y = 2^{13/16}, \quad z = 2^{-1/6}
    \]

    We need to find $\left|\log_2(x^4 y^3 z^2)\right|$:

    \[
    4 \log_2 x + 3 \log_2 y + 2 \log_2 z = 4 \cdot \frac{61}{48} + 3 \cdot \frac{13}{16} + 2 \cdot \left(-\frac{1}{6}\right)
    \]

    Converting to a common denominator (48):

    \[
    4 \cdot \frac{61}{48} = \frac{244}{48}, \quad 3 \cdot \frac{13}{16} = \frac{117}{48}, \quad 2 \cdot \left(-\frac{1}{6}\right) = -\frac{16}{48}
    \]

    Summing these:

    \[
    \frac{244}{48} + \frac{117}{48} - \frac{16}{48} = \frac{345}{48} = \frac{115}{16}
    \]

    The fraction $\frac{115}{16}$ is in simplest form, so $m = 115$ and $n = 16$. Therefore, $m + n = 115 + 16 = 131$.

    \[
    \boxed{131}
    \]
    \end{quote}
}

\testcase{
    \begin{quote}
    ``\textbar User \textbar Real numbers $x$ and $y$ with $x,y>1$ satisfy \(\log_x(y^x)=\log_y(x^{4y})=10.\) 
    What is the value of \(xy\)?\newline
    Please reason step by step, and put your final answer within \(\boxed{}\). \textbar Assistant\textbar''
    \end{quote}%
}{
    \begin{quote}
    $\langle$think$\rangle$...$\langle$/think$\rangle$ (15147 tokens)\newline

    Given the equations:\newline

    1. \(\log_x(y^x) = 10\) \newline
    2. \(\log_y(x^{4y}) = 10\) \newline

    We start by applying the logarithm power rule to both equations.\newline

    For the first equation:\newline
    \[
    \log_x(y^x) = x \cdot \log_x(y) = 10
    \]
    Let \(z = \log_y(x)\). Then, \(\log_x(y) = \frac{1}{z}\). Substituting this into the equation, we get:\newline
    \[
    x \cdot \frac{1}{z} = 10 \implies x = 10z
    \]

    For the second equation:\newline
    \[
    \log_y(x^{4y}) = y \cdot \log_y(x^4) = 10
    \]
    Using the power rule, we get:\newline
    \[
    y \cdot 4 \log_y(x) = 10 \implies 4y \log_y(x) = 10
    \]
    Since \(\log_y(x) = z\), this simplifies to:\newline
    \[
    4yz = 10 \implies yz = \frac{5}{2}
    \]

    We now have two expressions:\newline
    1. \(x = 10z\)\newline
    2. \(y = \frac{5/2}{z}\)\newline

    Substituting \(x = 10z\) and \(y = \frac{5/2}{z}\) into the product \(xy\), we get:\newline
    \[
    xy = (10z) \cdot \left(\frac{5/2}{z}\right) = 10z \cdot \frac{5}{2z} = 10 \cdot \frac{5}{2} = 25
    \]

    Thus, the value of \(xy\) is \(\boxed{25}\).
    \end{quote}%
}{
    \begin{quote}
    $\langle$think$\rangle$...$\langle$/think$\rangle$ (16223 tokens)\newline

    Given the problem where real numbers \(x\) and \(y\) with \(x, y > 1\) satisfy the equations \(\log_x(y^x) = 10\) and \(\log_y(x^4y) = 10\), we need to find the value of \(xy\).\newline

    First, we convert the logarithmic equations to their exponential forms using the change of base formula.\newline

    1. \(\log_x(y^x) = 10\)\newline
       \[
       x \log_x(y) = 10
       \]
       Using the change of base formula, \(\log_x(y) = \frac{\ln y}{\ln x}\). Since \(\log_y x = \frac{9}{4}\), we have \(\log_x y = \frac{4}{9}\). Thus,\newline
       \[
       x \cdot \frac{4}{9} = 10 \implies x = \frac{10 \cdot 9}{4} = \frac{90}{4} = \frac{45}{2}
       \]

    2. \(\log_y(x^4y) = 10\)\newline
       \[
       4 \log_y x + 1 = 10 \implies 4 \log_y x = 9 \implies \log_y x = \frac{9}{4}
       \]
       Again, using the change of base formula, \(\log_x y = \frac{4}{9}\).\newline

    Next, we express \(x\) and \(y\) in terms of each other:\newline
       \[
       x = \frac{45}{2} \quad \text{and} \quad y = \left(\frac{45}{2}\right)^{4/9}
       \]

    We then compute the product \(xy\):\newline
       \[
       xy = \left(\frac{45}{2}\right) \cdot \left(\frac{45}{2}\right)^{4/9} = \left(\frac{45}{2}\right)^{1 + 4/9} = \left(\frac{45}{2}\right)^{13/9}
       \]

    However, upon re-evaluating and using another approach by setting \(x = 2^a\) and \(y = 2^b\), we find:\newline
       \[
       \log_x(y^x) = 10 \implies \frac{x \log_x y}{\log_x y} = 10 \implies x \cdot \frac{4}{9} = 10 \implies x = \frac{45}{2}
       \]
       \[
       \log_y(x^4y) = 10 \implies \frac{4 \log_y x + 1}{\log_y x} = 10 \implies 4 \cdot \frac{9}{4} + 1 = 10 \implies 9 + 1 = 10 \quad \text{consistent}
       \]

    Finally, we find that the product \(xy\) simplifies to:\newline
       \[
       xy = \left(\frac{45}{2}\right)
       \]
    \end{quote}%
}

\end{CJK*}
\end{document}